\documentclass[10pt]{article} 

\usepackage[accepted]{rlj} 

\usepackage{amssymb}            
\usepackage{mathtools}          
\usepackage{mathrsfs}           
\usepackage{eucal}[mathscr]
\usepackage{graphicx}           
\usepackage{subcaption}         
\usepackage[space]{grffile}     
\usepackage{url}                
\usepackage{lipsum}             
\usepackage{bm}
\usepackage{bbm}
\usepackage{algorithm}
\usepackage{algpseudocode}
\usepackage{float}
\usepackage[skip=0pt]{caption}

\DeclareMathOperator*{\argmax}{argmax}

\title{A Study of Value-Aware Eigenoptions}

\setrunningtitle{A Study of Value-Aware Eigenoptions}

\author{Harshil Kotamreddy\textsuperscript{1,2}, Marlos C. Machado\textsuperscript{1,2,3}}

\emails{\{kotamred,machado\}@ualberta.ca}

\affiliations{
$^{1}$\textbf{Department of Computing Science, University of Alberta, Canada}\\
$^{2}$\textbf{Alberta Machine Intelligence Institute (Amii)}\\
$^{3}$\textbf{CIFAR AI Chair}\\
}

\contribution{
    Provide a succinct but precise list of the contribution(s) of the paper. Use contextual notes to avoid implications of contributions more significant than intended and to clarify and situate the contribution relative to prior work (see the examples below). If there is no additional context, enter ``None''. Try to keep each contribution to a single sentence, although multiple sentences are allowed when necessary. If using complete sentences, include punctuation. If using a single sentence fragment, you may omit the concluding period. A single contribution can be sufficient, and there is no limit on the number of contributions. Submissions will be judged mostly on the contributions claimed on their cover pages and the evidence provided to support them. Major contributions should not be claimed in the main text if they do not appear on the cover page. Overclaiming can lead to a submission being rejected, so it is important to have well-scoped contribution statements on the cover page.
    }
    {
    None
    }

\contribution{
    The submission template for submissions to RLJ/RLC 2025
    }
    {
    Built from previous RLC/RLJ, ICLR, and TMLR submission templates
    }

\contribution{
    \textit{{[}Example of one contribution and corresponding contextual note for the paper ``Policy gradient methods for reinforcement learning with function approximation'' \citep{Sutton2000}.]}\\ This paper presents an expression for the policy gradient when using function approximation to represent the action-value function.
    }
    {
    Prior work established expressions for the policy gradient without function approximation \citep{Williams1992}.
    }


\summary{The summary appears on the cover page. Although it can be identical to the abstract, it does not have to be. One might choose to omit the stated contributions in the Summary, given that they will be stated in the box below. The original abstract may also be extended to two paragraphs. The authors should ensure that the contents of the cover page fit entirely on a single page. The cover page does \textbf{not} count towards the 8--12 page limit.

\lipsum[1]
}

\begin{document}

\maketitle  

\begin{abstract}
Options, which impose an inductive bias toward temporal and hierarchical structure, offer a powerful framework for reinforcement learning (RL). While effective in sequential decision-making, they are often handcrafted rather than learned. Among approaches for discovering options, eigenoptions have shown strong performance in exploration, but their role in credit assignment remains underexplored. In this paper, we investigate whether eigenoptions can accelerate credit assignment in model-free RL, evaluating them in tabular and pixel-based gridworlds. We find that pre-specified eigenoptions aid not only exploration but also credit assignment, whereas online discovery can bias the agent’s experience too strongly and hinder learning. In the context of deep RL, we also propose a method for learning option-values under non-linear function approximation, highlighting the impact of termination conditions on performance. Our findings reveal both the promise and complexity of using eigenoptions, and options more broadly, to simultaneously support credit assignment and exploration in reinforcement learning. 
\end{abstract}

\section{Introduction}

While reinforcement learning (RL) has achieved many successes recently~\citep[e.g.,][]{Silver2016,silver2018general,openai2019dota2largescale,Degrave22,Ouyang22}, one arguably underexplored aspect is the effective use of temporal abstractions, such as options~\citep{SUTTON1999181}. These abstractions introduce an inductive bias that departs from the dominant end-to-end learning paradigm. Notably, in several high-profile cases, such abstractions were handcrafted by human experts in advance and proved instrumental to the system's success~\citep[e.g.,][]{Vinyals19,Bellemare20}.\looseness=-1

Autonomously discovering temporal abstractions, rather than relying on handcrafted ones, remains a rich and active area of research~\citep[e.g.,][]{feudalRL,skillchaining,optioncritic,vezhnevets2017feudal,deliberation,deepskillchaining,eysenbachdiversity}. Importantly, this diversity of approaches reflects the broad range of challenges where temporal abstractions have been applied, including credit assignment~\citep{SUTTON1999181,Solway14}, exploration~\citep{Jong08,pmlr-v97-jinnai19b}, and generalization~\citep{Konidaris2007BuildingPO}. \looseness=-1

Eigenoptions are a specific class of options originally introduced to address exploration challenges~\citep{machado2016learning,machado2017laplacian}, and have since been shown to be scalable and to lead to state-of-the-art performance in a range of high-dimensional problems~\citep{klissarov2023deep}. Yet, despite being defined in terms of how information diffuses through the environment, a property closely related to temporal credit assignment, few efforts have explored eigenoptions for that purpose. The limited work that applied eigenoptions beyond exploration did not explicitly investigate their potential for aiding credit assignment~\citep{liu2017eigenoption,sutton2023reward}. \looseness=-1

In this paper, we explicitly investigate whether eigenoptions can be used to accelerate credit assignment in model-free RL. We consider scenarios in which options are provided in advance and those in which they are discovered online, evaluating their impact across both tabular and pixel-based gridworlds. Our findings show that when options are available beforehand, they can aid not only exploration but also credit assignment, extending the list of benefits afforded by eigenoptions.

However, when options are discovered online and simultaneously used for exploration, their influence on the agent's behavior becomes more pronounced. Early inaccuracies in option learning can skew the agent's experience, limiting its ability to fully explore the environment, thereby hindering credit assignment. This highlights a previously underexplored interaction between exploration and credit assignment in the context of options. Finally, we propose a method for learning option-values under non-linear function approximation, where defining effective termination conditions is particularly challenging due to approximation errors. Collectively, these results underscore both the potential and the complexity of using temporal abstractions for credit assignment in RL. \looseness=-1

\section{Preliminaries}

In this paper, we use capital letters to designate random variables, calligraphic font for sets, lowercase bold letters for vectors, and uppercase bold letters to denote matrices. $\Delta(\mathcal{X})$ is the set of probability distributions over set $\mathcal{X}$. \looseness=-1

\subsection{Reinforcement Learning}
In the reinforcement learning setting, an agent interacts with an environment with the aim of maximizing reward. RL problems are typically represented by a Markov Decision Process (MDP). An MDP is defined as a tuple $\langle\mathcal{S}, \mathcal{A}, p, r\rangle$ where $\mathcal{S}$ is the set of all states, $\mathcal{A}$ is the set of all actions, $p(s| s', a) = \mathbb{P}[S_{t + 1} = s' | S_t = s, A_t = a]$ is the transition probability kernel, and $r(s, a) = \mathbb{E}[R_{t+1} | S_t = s, A_t = a]$ is the expected reward given the agent takes action $a$ in state $s$. At every time step $t$, the agent is in state $S_t$ and interacts with the environment by taking action $A_t$; then the agent moves to state $S_{t + 1}$ according to $p$ and receives a reward $R_{t+1}$.

The agent learns a policy $\pi: \mathcal{S} \rightarrow \Delta(\mathcal{A})$ that maximizes the expected discounted return, $G_t = \sum_{k=0}^{\infty} \gamma^k R_{t + k + 1}$, where $\gamma \in \left[0, 1\right)$ is the discount factor. We focus on value-based methods that estimate the state-action value function $q_\pi(s, a) = \mathbb{E}_\pi[G_t|S_t \! = \! s, A_t \! = \! a]$ for the optimal policy, $\pi^*$. Specifically, we use Q-learning \citep{Watkins1992} which estimates $q_{\pi^*}$ using the update rule \looseness=-1
\vspace{-1pt}
\begin{equation}
    Q(S_t, A_t) \leftarrow Q(S_t, A_t) + \alpha \left(R_{t + 1} + \gamma \max_{a \in \mathcal{A}} Q(S_{t+1}, a) - Q(S_t, A_t)\right),
\end{equation}
where $\alpha$ is the step size. The policy is defined as $\pi(s) = \argmax_{a \in \mathcal{A}} Q(s, a).$

\subsection{Options}
Options in RL are a framework for temporal abstraction, letting agents act over multiple time steps \citep{SUTTON1999181}. An option is defined as a tuple, $o = \langle \mathcal{I}_o, \pi_o, \beta_o\rangle$, where $\mathcal{I}_o \subseteq \mathcal{S}$ is the option's initiation set, $\pi_o: \mathcal{S} \rightarrow \Delta(\mathcal{A}) $ is the option's policy, and $\beta_o: \mathcal{S} \rightarrow [0, 1]$ is the option's termination function. An option can be taken in any state within the initiation set $\mathcal{I}_o$, after which it acts according to the option policy $\pi_o$ until it terminates according to $\beta_o$. Note that actions in $\mathcal{A}$ are one-step options.

We can learn state-option estimates, that is, the expected return if the agent takes option $O_t$ in state $S_t$, using the SMDP Q-Learning update rule upon option termination \citep{SUTTON1999181}: \looseness=-1
\vspace{-1pt}
\begin{equation}
    Q(S_t, O_t) \leftarrow Q(S_t, O_t) + \alpha \left[R + \gamma^K \max_{o \in \mathcal O_{S_{t+K}}} Q(S_{t+K}, o) - Q(S_t, O_t)\right],
\end{equation}
where $o \in \mathcal{O}$ is an option, $\mathcal{O}$ is the set of all options, $Q(s, o)$ is the estimate of the optimal state-option value function, $\alpha$ is the step size, $K$ is the number of time steps the option took before termination, $R$ represents the cumulative discounted reward through the duration of the option, and $\mathcal{O}_S$ is the set of all options that contain state $S$ within their initiation set, $\mathcal{I}_o$.

Intra-option Q-learning \citep{sutton1998intra} improves on SMDP Q-learning by also updating $Q(s, o)$ while the agent is still acting according to an option's policy, in contrast to applying the update rule only upon an option's termination:
\vspace{-1pt}
\begin{equation} \label{eqn: intra-option}
    Q(S_t, O_t) \leftarrow Q(S_t, O_t) + \alpha \left[ \left( R_{t+1} + \gamma U\left(S_{t+1}, O_t\right) \right) - Q(S_t, O_t) \right],
\end{equation}
where $U(s, o) = \left(1 - \beta_o(s)\right)Q(s, o) + \beta_o(s) \max_{o' \in \mathcal{O}_s} Q(s, o').$ This allows us to perform state-option updates to all options if their policies take the same action as the current option and/or action. \looseness=-1

\subsection{Eigenoptions} 
Eigenoptions~\citep{machado2017laplacian, machado2018eigenoption} are options discovered through the eigenvectors of the successor representation (SR)~\citep{dayan1993improving}. The SR encodes the ``temporal distance'' between states. It represents each state $s$ as an $|\mathcal{S}|$ dimensional vector that contains the expected discounted visitation from $s$ to every state $s' \in \mathcal{S}$. The SR matrix, $\mathbf{\Psi_\pi}$, with respect to a policy $\pi$, can be calculated using $\mathbf{P_\pi}$, the transition probability matrix induced by $\pi$:
\vspace{-1pt}
\begin{equation} \label{eqn:SR_closed}
    \mathbf{\Psi_\pi} = \left( \mathbf{I} - \gamma \mathbf{P_\pi} \right) ^{-1},
\end{equation}
where $\mathbf{I}$ is the identity matrix. When $\mathbf{P_\pi}$ is unknown, given a step size $\eta$, the SR can be estimated from samples using the temporal difference update rule for state $S_t$ at time step $t$:
\vspace{-1pt}
\begin{equation} \label{eqn:onlineSR}
    \hat{\Psi}\left(S_t, j\right) \leftarrow \hat{\Psi}\left(S_t, j\right) + \eta \left[\mathbbm{1}_{\{S_t = j\}} + \gamma \hat{\Psi}\left(S_{t+1}, j\right) - \hat{\Psi}\left(S_t, j\right) \right] \quad \forall j \in \mathcal{S}.
\end{equation} 

Eigenoptions are options that maximize an intrinsic reward function defined by the eigenvectors of the SR.
The intrinsic reward function used to generate an option policy using eigenvector $\bm{e}$ of $\boldsymbol{\Psi}_\pi$ is: \looseness=-1
\begin{align} \label{eqn:intrinsic_reward}
    r^{\bm{e}}(s, s') = \bm{e}^\top \left( \boldsymbol{\phi}(s') - \boldsymbol{\phi}(s)\right) \quad \forall s, s' \in \mathcal{S},
\end{align}
where $\boldsymbol{\phi}(s)$ is the feature representation of state $s$. 

Eigenoptions can also be discovered online through the Representation-Driven Option Discovery (ROD) cycle \citep{machado2023temporal,Machado25}. The cycle starts with an RL agent gathering data by interacting with its environment and then using this data to learn a representation of the environment. In this case, the agent learns the SR and its eigenvectors. The agent then uses the learned representation to create an intrinsic reward function (see Eq. \ref{eqn:intrinsic_reward}), and then learns an option policy to maximize it. Finally, the option's initiation set is defined as all states with positive Q-values, and the complement is defined as termination states. This can be done because the intrinsic reward function is naturally defined such that the agent receives negative rewards when it exits what should be a termination state. \looseness=-1

Once an option is created, it is added to the agent's option set, and the cycle repeats. So far, such options have only been used for exploration. Specifically, in an $\epsilon$-greedy step, the agent might sample an option and act according to its policy until termination. Through this process, the agent explores areas of the environment that are less frequently visited, enabling much faster and more efficient exploration. This algorithm is called covering eigenoptions (CEO)~\citep{machado2023temporal}.
\vspace{-3pt}
\section{Learning Option-Values for Eigenoptions in the Tabular Setting} \label{sec:tabular}
In the tabular case, eigenoptions defined using the first $n$ eigenvectors of the SR have been shown to significantly reduce the number of time steps required to visit all states of an environment \citep{machado2023temporal}. Additionally, perhaps because they capture information on how rewards diffuse in the environment, they have also been shown to be quite effective when used for planning~\citep{sutton2023reward}. Thus, a very natural question in the model-free setting is whether we should see even faster learning if we were to learn option-values for eigenoptions; using them not only for exploration but also for credit assignment. We call this approach value-aware eigenoptions (VAEO). We consider bottleneck options as a baseline because they are the most traditional approach for credit assignment~\citep{mcgovern2001accelerating, simsekbetweenness} and they can seen as forming an optimal behavioral hierarchy~\citep{Solway14}.

\begin{figure}[t]
    \centering
    \includegraphics[width=0.87\linewidth]{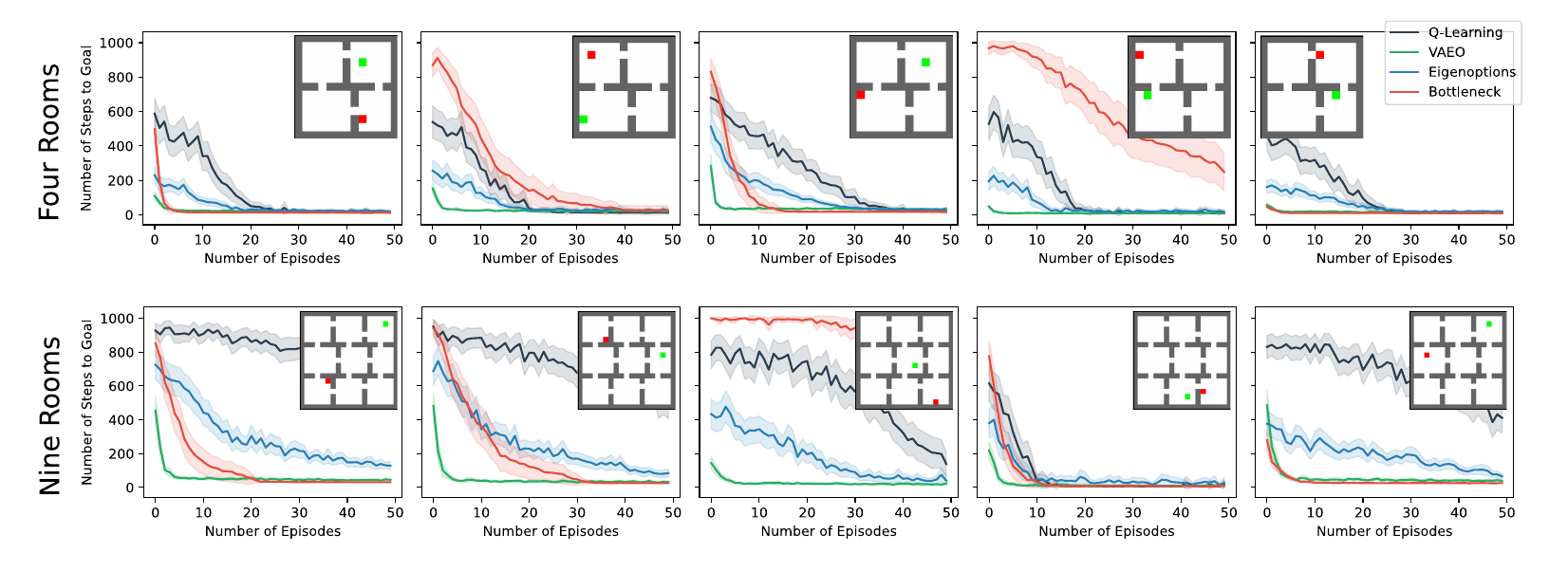}
    \vspace{-1pt}
    \caption{Performance of value-aware eigenoptions (VAEO) compared to baselines. VAEO outperforms the other algorithms in most configurations. Environment configurations are inset in each plot. Green squares represent start states, and red squares represent goal states. We use six and 24 eigenoptions in the four rooms and nine rooms domains, respectively. We use all bottleneck options, which totals to eight in the four rooms domain and 24 in the nine rooms domain. We evaluate all algorithms for 100 independent runs and the shaded region represents a $99\%$ confidence interval.}
    \label{fig:tabVAEO}
\end{figure}

\vspace{-1mm}
\subsection{Acting and Learning with Eigenoptions} 

We first focus on the benefits of value-aware eigenoptions without confounding factors from the option discovery process. Thus, we first consider options obtained through the closed form of the SR (Eq. \ref{eqn:SR_closed}). We demonstrate that in the tabular setting, learning option-values for pre-computed eigenoptions leads to faster learning when compared to using pre-computed eigenoptions strictly for exploration. We perform experiments in the four rooms and nine rooms domains with randomly generated start and goal state configurations (inset in Figure \ref{fig:tabVAEO}). We use a modified version of the Minigrid environment \citep{MinigridMiniworld23} with cardinal actions. The agent receives zero reward at every transition except those that lead to the goal state, which lead to +1 reward. \looseness=-1

We generate eigenoptions using the top $n$ eigenvectors of the SR. We learn the option policies using Q-learning with samples from random exploration of the environment, where the agent starts at a random location every episode. Episodes are 1000 steps long, and in this phase, we do not treat goal states as absorbing states but as regular states. To learn option-values for these precomputed eigenoptions, we use intra-option Q-learning. If an action $a$ is taken in a state $s$, regardless of how it was selected, we update all options that would have taken action $a$ in state $s$.

We report the number of time steps it takes the agent to reach the goal state during every episode. We compare the performance of Value-Aware Eigenoptions (VAEO) to Q-learning with eigenoptions used for exploration~\citep{machado2018eigenoption, machado2023temporal} and intra-option Q-learning with bottleneck options~\citep{sutton1998intra}. We did a hyperparameter sweep to decide on the number of eigenoptions. \looseness=-1

As shown in Figure \ref{fig:tabVAEO}, VAEO outperforms eigenoptions and bottleneck options in nearly all configurations of the environment. This suggests that credit assignment provides additional benefit on top of the exploration bonus provided by eigenoptions. As initially predicted, learning option-values allows the agent to exploit options that it previously found to be useful to maximize its return.

\begin{figure}[t]
    \centering
    \includegraphics[width=0.87\linewidth]{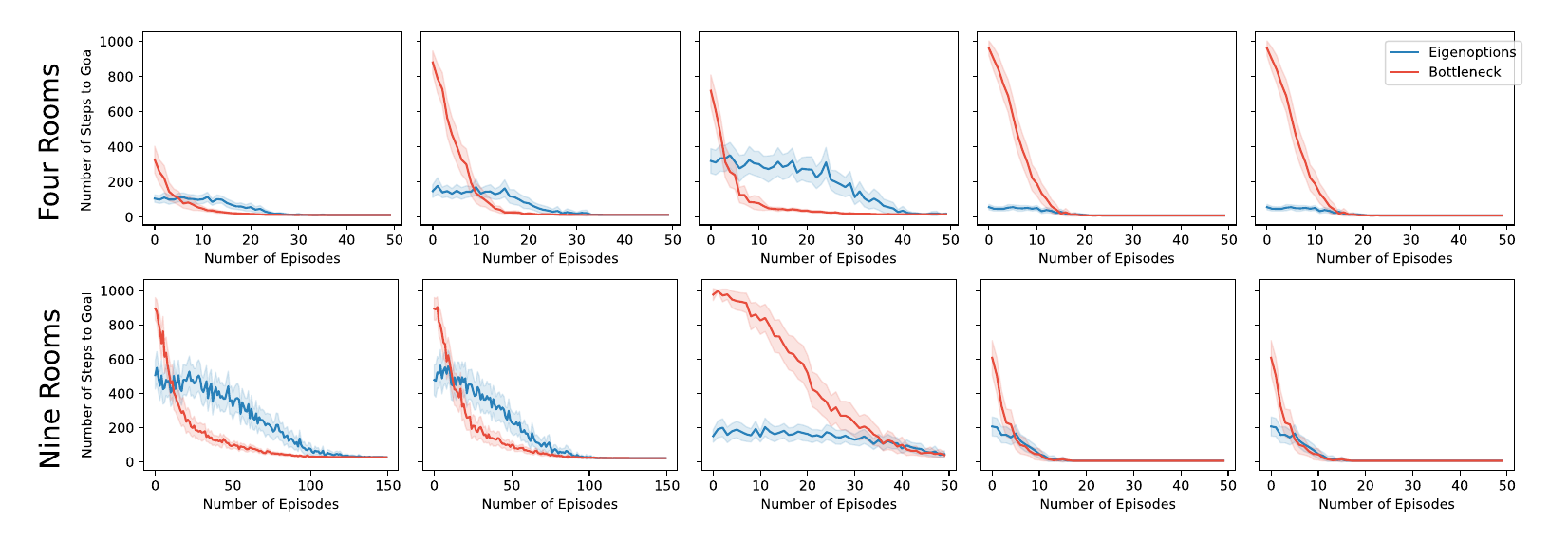}
    \caption{Credit assignment through an evaluation phase. We evaluate both algorithms for 100 seeds and report the 99\% confidence interval. Environment configurations are the same as in Fig.~\ref{fig:tabVAEO}.}
    \label{fig:tabCredit}
    \vspace{-2mm}
\end{figure}

\vspace{-1mm}
\subsection{Credit Assignment}
While these results seem to indicate that credit assignment through options improves performance, with intra-option Q-learning we are essentially treating options as additional actions. Thus, the results above still confound the benefits of using eigenoptions for credit assignment and exploration.

To factor out the impact of the exploration benefits induced by eigenoptions, we constrained the agent to select only primitive actions while performing an intra-option update at every time step to learn option-values through the agent's actions. This eliminates any and all exploration benefits the agent gets from acting according to an option's policy. At the end of every episode, we evaluate the agent's time-to-goal when also considering options in its action space. In this case, the performance we see is strictly due to credit being assigned to the eigenoptions.

As shown in Figure \ref{fig:tabCredit}, eigenoptions initially assign credit much more quickly than bottleneck options. However, they can stay at suboptimal trajectories longer. These results are not too surprising. Bottleneck options tend to be shorter, and the solution to any start/goal configuration unavoidably leads the agent through a bottleneck state, making bottleneck options part of the optimal policy~\citep{Solway14, sutton2023reward}. In fact, it is somewhat surprising that eigenoptions are so competitive in many of these settings. We conjecture that this might be due to how eigenoptions are obtained through the environment's diffusive information flow, and to how   different eigenoptions operate at different time scales, potentially allowing some of them to be a part of the optimal policy.

\section{Challenges when Discovering Options Online} \label{sec:tabularROD}
We now discuss learning option-values when options are discovered online. We introduce a value-aware version of CEO called value-aware covering eigenoptions (VACE). In VACE, every time a new option is added to the option set, we initialize an additional vector in our Q-function to store option-values for the newly created option. As the agent interacts with the environment, we use intra-option Q-learning updates to learn option-values. The VACE algorithm is detailed in Appendix \ref{appendix:VACE}.

To see how much of a benefit value-aware eigenoptions provide when eigenoptions are discovered online, we compare the performance of VACE to CEO and Q-learning, with the latter using $\epsilon$-greedy exploration. The experiment setup is the same as when using precomputed options. We evaluate these algorithms when discovering new options every 1000 timesteps.

\begin{figure}[t]
    \centering
    \includegraphics[width=0.87\linewidth]{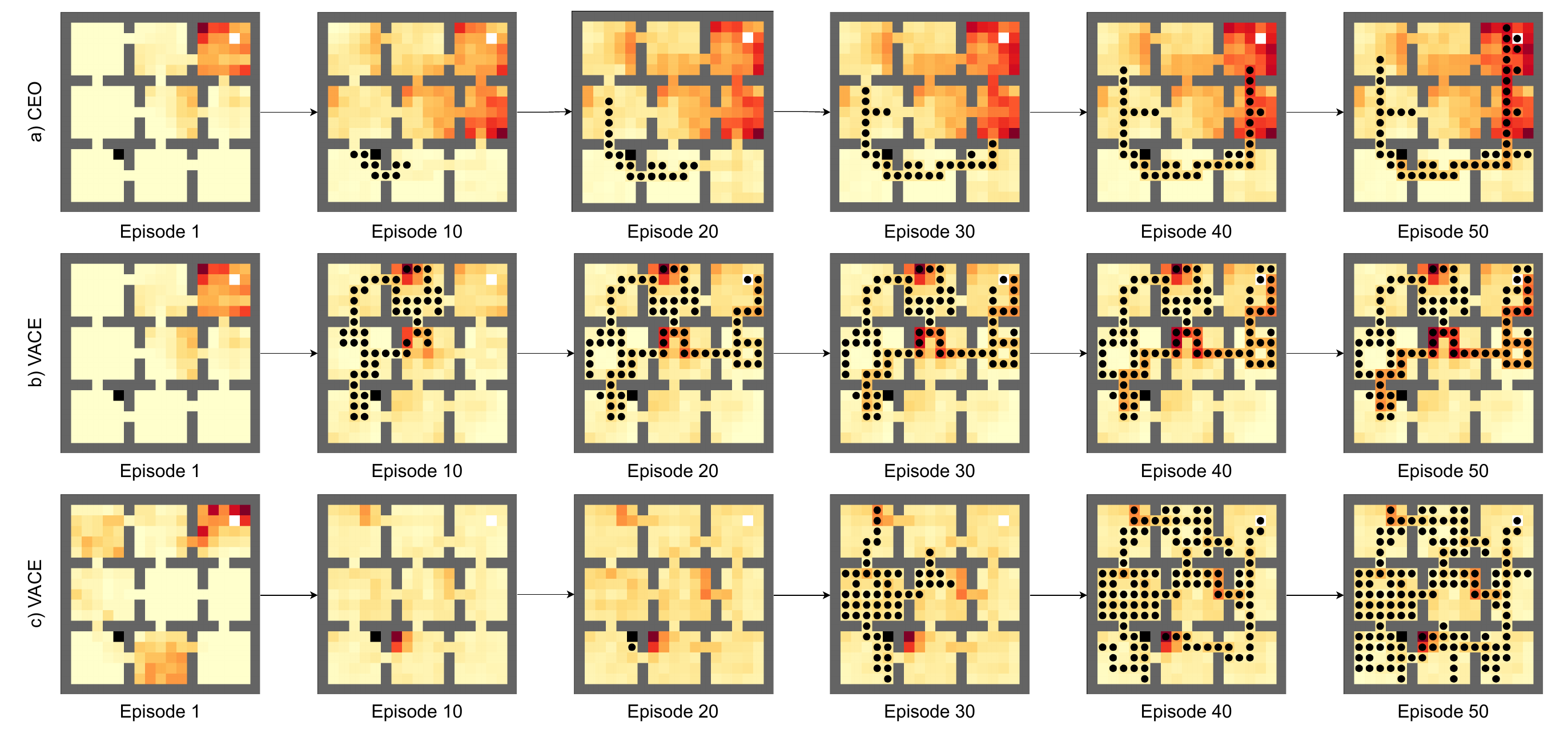}
    \caption{Value propagation in a) a typical run of CEO, b) a typical run of VACE and c) a bad run of VACE. The start state is highlighted in white, while the goal state is highlighted in black. Black dots represent states with a value greater than zero. Darker red states represent states frequently visited, while lighter yellow squares represent states visited infrequently.}
    \label{fig:valueprop}
\end{figure}

As shown in Fig. \ref{fig:ROD} in Appendix \ref{appendix:VACE}, VACE outperforms CEO in the four rooms environment while the performance gain in the nine rooms environment is much less apparent. VACE still outperforms CEO in median performance, though (see Fig.~\ref{fig:RODMedian} in Appendix \ref{appendix:VACE}). This is because VACE outperforms CEO on most runs, but it performs significantly worse in a few runs. When options are used for both credit assignment and exploration, inaccurate estimates of an option's value might make a suboptimal option be selected much more often than it should be in the argmax. Such an option has a much more persistent behaviour, with long-term consequences, when compared to a wrongly selected primitive action. Thus, in VACE, the discovered options do help the agent visit underexplored regions in the environment, but because the discovered options are treated as actions, they have a much bigger impact on the agent, sometimes making it harder for the agent to find the goal state. 

Let us elaborate. Consistently acting according to option's policies can sometimes hinder performance. In CEO, value propagation is relatively slow as value is propagated one state at a time, but the values start to be propagated relatively quickly because of effective exploration (see Fig.~\ref{fig:valueprop}a). Fig.~\ref{fig:valueprop}b shows a typical run of VACE. VACE propagates value much faster and, in this particular run, is able to create a value trace to the goal by episode 20. Fig.~\ref{fig:valueprop}c shows a bad run of VACE where it struggles to find the goal because the agent hops from one option to another, and even though it initially gets close to the goal state, it does not use enough random actions to get to the goal state because another option ``takes over'' before that. Thus, it takes much longer to start propagating values back. \looseness=-1

\section{Approximating Option-Values with Non-Linear Function Approximation}
Since value-aware eigenoptions lead to faster learning in the tabular setting, it is natural to ask whether they would also help in the function approximation setting. We investigate this question in the four rooms and nine rooms environments with pixel inputs. We again use a modified version of the Minigrid environment~\citep{MinigridMiniworld23} with cardinal actions. We first focus on using neural networks to approximate option-values, using tabular methods to compute the eigenvectors of the SR and the options themselves. Solutions to approximating the eigenvectors of the SR \citep{wulaplacian, wang2021towards, gomezproper} and the eigenoptions themselves \citep{klissarov2023deep} have already been discussed in the literature. We discuss nuances around option termination when using them for credit assignment in Section \ref{sec:approx}.

In all experiments, we use a two-layer convolutional network with 32 channels each, a 3 $\times$ 3 kernel, and a stride of 2, followed by a fully connected layer with 256 nodes and then the output layer with 4 nodes (one for each action) in the action-value network and 1 node in the option-value network. We implemented our algorithms with the PFRL library~\citep{pfrl}. \looseness=-1

\subsection{Tabular Option Policies} \label{sec:FAtabular}
To learn option-values in the function approximation setting, we use a hierarchical DQN architecture~\citep{kulkarni2016hierarchical} similar to that of \cite{deepskillchaining}. We use a separate neural network to learn each of the option-value functions on top of the value function for primitive actions (see Fig. \ref{fig:arch} in Appendix \ref{appendix:DVAEO}). 
The network parameters are represented as $\boldsymbol{\theta}^{o_k} \in \boldsymbol{\theta}^{o}$, where $k = 0$ represents the network that learns primitive action-values, and $k > 0$ represents the network that learns the option-values for option $k$. The target networks are represented as $\boldsymbol{\theta}^{o_{k}^{-}} \in \boldsymbol{\theta}^{o^-}$. The action or option with the highest value is selected during a greedy step. We use the DQN~\citep{Mnih2015} loss function $L_i = \mathbb{E}_{s, a \sim p(\cdot)} \left[\left(y_i - Q(s, a; \theta_i)\right)^2\right]$ with a modified Double DQN (DDQN)~\citep{van2016deep} target: \looseness=-1
\begin{equation}
    y_i = \mathbb{E}\left[r + \gamma U(s', o; \boldsymbol{\theta}^{o}_i, \boldsymbol{\theta}^{o^-}_{i}) \Big{|} s, a\right],
\end{equation}
where
$U(s', o; \boldsymbol{\theta}_i^o, \boldsymbol{\theta}^{o^-}_{i}) = \left( 1 - \beta_o(s)\right)Q(s, o; \boldsymbol{\theta}^{o^-}_i) + \beta_o(s)Q\left(s, \argmax_{o'}Q\left(s, o'; \boldsymbol{\theta}^{o}_i\right); \boldsymbol{\theta}^{o^-}_i\right).$

Similar to the experiments we performed in the tabular setting, we evaluate deep value-aware eigenoptions (DVAEO) against deep eigenoptions (DDQN with eigenoptions for exploration) and DDQN with $\epsilon$-greedy exploration. We use the same number of eigenoptions from the tabular setting, 6 eigenoptions in the four rooms domain and 24 eigenoptions in the nine rooms domain, for both DVAEO and deep eigenoptions. This hyperparameter was not swept in this setting. We populate the replay buffer with 1000 random transitions in the environment before starting the learning process.

\begin{figure}[t]
    \centering
    \includegraphics[width=0.87\linewidth]{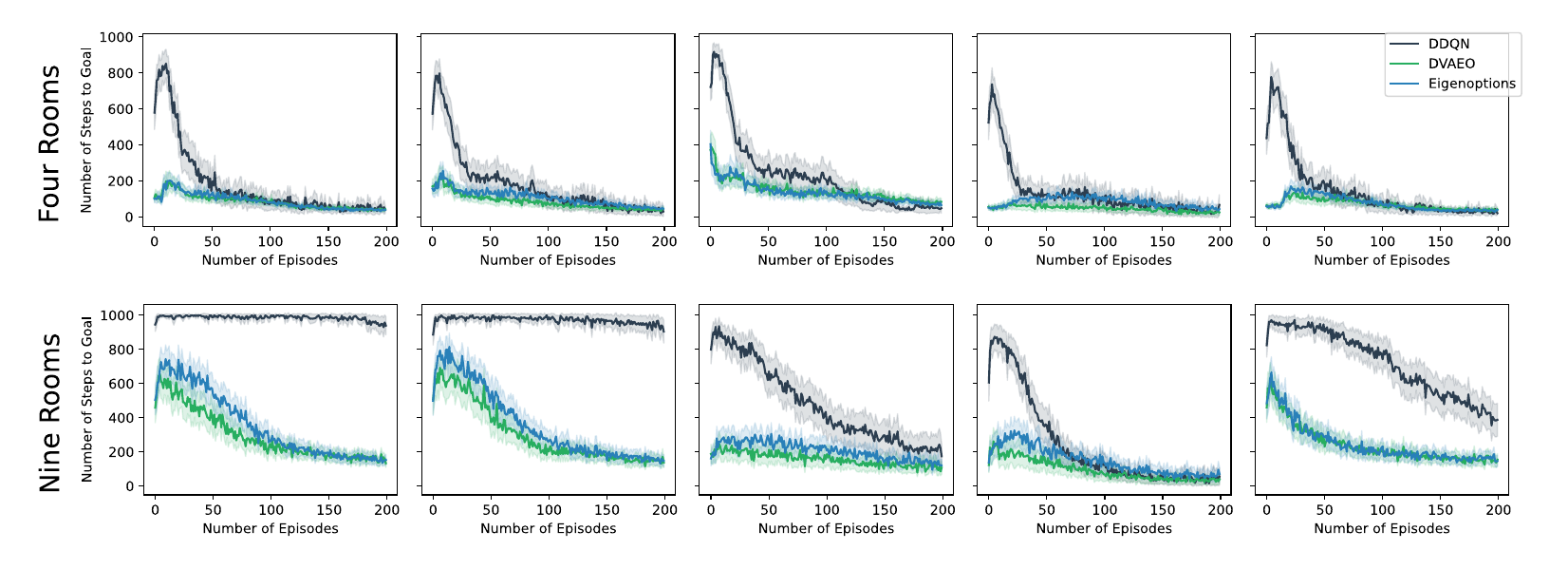}
    \caption{Performance of DVAEO and baselines with pixel observations. Environment configurations are the same as in Figure \ref{fig:tabVAEO}. We use 6 tabular eigenoptions in the four rooms domain and 24 tabular eigenoptions in the nine rooms domain for both DVAEO and deep eigenoptions. We train all algorithms for 100 independent runs and the shaded region represents a $99\%$ confidence interval.\looseness=-1}
    \label{fig:FATab}
\end{figure}

As shown in Figure \ref{fig:FATab}, DVAEO performs slightly better than deep eigenoptions in the nine rooms environment, while there is no difference in performance in the four rooms environment. The performance gains are not nearly as significant as those in the tabular setting, though. We hypothesize that this is because learning with neural networks is much slower than in the tabular case. This could imply that the exploration benefits from eigenoptions outweigh credit assignment in these experiments. However, the fact that there is a larger performance gain in the bigger environment might suggest that the benefits of DVAEO would become more evident in more complex environments.\looseness=-1

\subsection{Approximating Option Policies} \label{sec:approx}
We now extend the previous architecture by also approximating the option's policies and termination conditions. We use the DDQN algorithm to learn the option policies. 
Unlike the tabular setting, the method for determining whether to terminate an option at a given state is much more brittle because the generalization of neural networks leads to estimates of many states changing with a single update, making termination conditions based on specific thresholds ineffective. \cite{klissarov2023deep} successfully utilized option policies with nondeterministic termination using a constant $\beta = 0.1$. While this makes sense for exploration, given that randomized termination still leads to underexplored states, our initial experiments showed extremely high variance and slow learning when using a constant $\beta$. To reduce stochasticity, we terminate options 10 time steps after initiation; we obtained this value through a hyperparameter sweep over both environments we considered. \looseness=-1

As before, we evaluate DVAEO against deep eigenoptions (now with deep option policies) and DDQN with $\epsilon$-greedy exploration. The option policies were trained on a version of the environment without start or goal states for 500 episodes, ensuring that the learned policies matched the tabular policies. 
We again use 6 eigenoptions in the four rooms environment and 24 eigenoptions in the nine rooms environment for both DVAEO and deep eigenoptions, and we populate the replay buffer with 1000 random samples from the environment before starting the learning process. \looseness=-1

DVAEO with DDQN option policies performs similarly to deep eigenoptions in most configurations (see Fig.~\ref{fig:FA} in Appendix \ref{appendix:DVAEO}). Using eigenoptions for exploration led to similar performance as when using them for both exploration and credit assignment. This can be attributed to suboptimal termination conditions. The DDQN option policies were functionally the same as the tabular ones, except for the termination condition. When termination states are not well defined, there are some states where the option's policy leads the agent into a loop and/or a wall. Since the agent cannot terminate at such states, it has to continue taking the option until it terminates either through chance or after $n$ steps. This issue is exacerbated if there is poor network initialization or bad updates due to neural network generalization. Interrupting such options might be a promising avenue for future work. \looseness=-1

\vspace{-0.2cm}
\section{Related Work}
\vspace{-0.2cm}

Using options for credit assignment is an active area of research. A common approach involves identifying bottleneck states, critical states that must be traversed to reach large regions of the state space, and constructing options that lead to them \citep{mcgovern2001automatic, csimcsek2004using, simsekbetweenness}. This strategy improves credit assignment by shifting it from individual states to broader regions \citep{SUTTON1999181, Solway14}. \citet{kulkarni2016hierarchical} introduced options into the deep RL setting, inspiring a series of methods based on the hierarchical DQN framework. Additionally, skill chaining has been shown to accelerate credit assignment by chaining options along the path to the goal, allowing full option transitions to contribute to policy updates \citep{skillchaining, deepskillchaining}. Feudal methods \citep{vezhnevets2017feudal, levy2019learning} have also shown promise, using higher-level policies to assign credit over extended horizons. For a much deeper and comprehensive discussion on option discovery in general, we refer the reader to the recent survey written by \citet{Klissarov2025} on the topic.

The idea of using eigenoptions for credit assignment is not entirely new. \cite{liu2017eigenoption} used the eigenvectors of the SR as an auxiliary reward within the option-critic architecture. The options are learned using an intrinsic reward that combines the eigenvectors of the SR with environmental reward, making eigenoptions reward-respecting \citep{sutton2023reward}. While this approach has merit and significantly speeds up learning compared to option-critic~\citep{optioncritic}, we aimed to evaluate the utility of options defined strictly through the eigenvectors of the SR for credit assignment. \looseness=-1

Methods that use the Default Representation (DR) \citep{Piray2021} can also be viewed as incorporating environmental rewards into eigenoptions. The DR captures the expected reward between two states instead of the temporal distance between them, as the SR does, which gives rise to eigenoption variants obtained from reward-aware representations \citep{tse2025reward}. \looseness=-1

\vspace{-0.2cm}
\section{Conclusion}
\vspace{-0.2cm}

In this paper, we studied value-aware eigenoptions. We (1) demonstrated the benefits of learning option-values for pre-computed eigenoptions, (2) introduced VACE, a method that builds on CEO to learn option-values for eigenoptions learned online, and (3) introduced DVAEO, a deep RL method to learn option-values for eigenoptions. Maybe even more important than the methods themselves were the discussions around the impact of termination conditions in the function approximation setting and the challenges of learning option-values for eigenoptions that are discovered online.

Future work may involve methods that learn a well-defined termination set for eigenoptions in the function approximation setting. Potential approaches might involve thresholding \citep{jinnai2020exploration} and clustering \citep{ramesh2019successor} based on the learned value function. Additionally, developing techniques to maintain the exploration benefits of eigenoptions once they are added to an agent's action space will be crucial to using option-values effectively when eigenoptions are learned online.

More generally, this work outlines the potential and challenges associated with using eigenoptions for credit assignment in model-free RL. We have demonstrated that eigenoptions exhibit potential for credit assignment, despite being initially introduced for exploration. However, their potential for credit assignment seems to be overshadowed by interactions with the state visitation distribution they induce when they are learned online, or when used in the function approximation setting.

\section*{Acknowledgements}

This research was supported in part by the Natural Sciences and Engineering Research Council of Canada (NSERC) and the Canada CIFAR AI Chair Program. Computational resources were provided in part by the Digital Research Alliance of Canada.

\appendix



\bibliography{main}

\begin{thebibliography}{50}
\providecommand{\natexlab}[1]{#1}
\providecommand{\url}[1]{\texttt{#1}}
\expandafter\ifx\csname urlstyle\endcsname\relax
  \providecommand{\doi}[1]{DOI: #1}\else
  \providecommand{\doi}{DOI: \begingroup \urlstyle{rm}\Url}\fi

\bibitem[Bacon et~al.(2017)Bacon, Harb, and Precup]{optioncritic}
Pierre-Luc Bacon, Jean Harb, and Doina Precup.
\newblock The option-critic architecture.
\newblock In \emph{AAAI Conference on Artificial Intelligence}, 2017.

\bibitem[Bagaria \& Konidaris(2019)Bagaria and Konidaris]{deepskillchaining}
Akhil Bagaria and George Konidaris.
\newblock Option discovery using deep skill chaining.
\newblock In \emph{International Conference on Learning Representations}, 2019.

\bibitem[Bellemare et~al.(2020)Bellemare, Candido, Castro, Gong, Machado, Moitra, Ponda, and Wang]{Bellemare20}
Marc~G. Bellemare, Salvatore Candido, Pablo~Samuel Castro, Jun Gong, Marlos~C. Machado, Subhodeep Moitra, Sameera~S. Ponda, and Ziyu Wang.
\newblock {Autonomous navigation of stratospheric balloons using reinforcement learning}.
\newblock \emph{Nature}, 588:\penalty0 77--82, 2020.

\bibitem[Berner et~al.(2019)Berner, Brockman, Chan, Cheung, Dębiak, Dennison, Farhi, Fischer, Hashme, Hesse, Józefowicz, Gray, Olsson, Pachocki, Petrov, d.~O.~Pinto, Raiman, Salimans, Schlatter, Schneider, Sidor, Sutskever, Tang, Wolski, and Zhang]{openai2019dota2largescale}
Christopher Berner, Greg Brockman, Brooke Chan, Vicki Cheung, Przemysław Dębiak, Christy Dennison, David Farhi, Quirin Fischer, Shariq Hashme, Chris Hesse, Rafal Józefowicz, Scott Gray, Catherine Olsson, Jakub Pachocki, Michael Petrov, Henrique~P. d.~O.~Pinto, Jonathan Raiman, Tim Salimans, Jeremy Schlatter, Jonas Schneider, Szymon Sidor, Ilya Sutskever, Jie Tang, Filip Wolski, and Susan Zhang.
\newblock Dota 2 with large scale deep reinforcement learning.
\newblock \emph{CoRR}, abs/1912.06680, 2019.

\bibitem[Brockman et~al.(2016)Brockman, Cheung, Pettersson, Schneider, Schulman, Tang, and Zaremba]{gym}
Greg Brockman, Vicki Cheung, Ludwig Pettersson, Jonas Schneider, John Schulman, Jie Tang, and Wojciech Zaremba.
\newblock {OpenAI Gym}.
\newblock \emph{CoRR}, abs/1606.01540, 2016.

\bibitem[Chevalier-Boisvert et~al.(2023)Chevalier-Boisvert, Dai, Towers, de~Lazcano, Willems, Lahlou, Pal, Castro, and Terry]{MinigridMiniworld23}
Maxime Chevalier-Boisvert, Bolun Dai, Mark Towers, Rodrigo de~Lazcano, Lucas Willems, Salem Lahlou, Suman Pal, Pablo~Samuel Castro, and Jordan Terry.
\newblock Minigrid \& {Miniworld}: {Modular} \& customizable reinforcement learning environments for goal-oriented tasks.
\newblock \emph{CoRR}, abs/2306.13831, 2023.

\bibitem[\c{S}im\c{s}ek \& Barto(2008)\c{S}im\c{s}ek and Barto]{simsekbetweenness}
\"{O}zg\"{u}r \c{S}im\c{s}ek and Andrew Barto.
\newblock Skill characterization based on betweenness.
\newblock In \emph{Neural Information Processing Systems}, 2008.

\bibitem[Dayan(1993)]{dayan1993improving}
Peter Dayan.
\newblock Improving generalization for temporal difference learning: The successor representation.
\newblock \emph{Neural Computation}, 5\penalty0 (4):\penalty0 613--624, 1993.

\bibitem[Dayan \& Hinton(1992)Dayan and Hinton]{feudalRL}
Peter Dayan and Geoffrey~E. Hinton.
\newblock Feudal reinforcement learning.
\newblock In \emph{Neural Information Processing Systems}, 1992.

\bibitem[Degrave et~al.(2022)Degrave, Felici, Buchli, Neunert, Tracey, Carpanese, Ewalds, Hafner, Abdolmaleki, de~Las~Casas, Donner, Fritz, Galperti, Huber, Keeling, Tsimpoukelli, Kay, Merle, Moret, Noury, Pesamosca, Pfau, Sauter, Sommariva, Coda, Duval, Fasoli, Kohli, Kavukcuoglu, Hassabis, and Riedmiller]{Degrave22}
Jonas Degrave, Federico Felici, Jonas Buchli, Michael Neunert, Brendan~D. Tracey, Francesco Carpanese, Timo Ewalds, Roland Hafner, Abbas Abdolmaleki, Diego de~Las~Casas, Craig Donner, Leslie Fritz, Cristian Galperti, Andrea Huber, James Keeling, Maria Tsimpoukelli, Jackie Kay, Antoine Merle, Jean{-}Marc Moret, Seb Noury, Federico Pesamosca, David Pfau, Olivier Sauter, Cristian Sommariva, Stefano Coda, Basil Duval, Ambrogio Fasoli, Pushmeet Kohli, Koray Kavukcuoglu, Demis Hassabis, and Martin~A. Riedmiller.
\newblock Magnetic control of tokamak plasmas through deep reinforcement learning.
\newblock \emph{Nature}, 602\penalty0 (7897):\penalty0 414--419, 2022.

\bibitem[Eysenbach et~al.(2019)Eysenbach, Gupta, Ibarz, and Levine]{eysenbachdiversity}
Benjamin Eysenbach, Abhishek Gupta, Julian Ibarz, and Sergey Levine.
\newblock Diversity is all you need: Learning skills without a reward function.
\newblock In \emph{International Conference on Learning Representations}, 2019.

\bibitem[Fujita et~al.(2021)Fujita, Nagarajan, Kataoka, and Ishikawa]{pfrl}
Yasuhiro Fujita, Prabhat Nagarajan, Toshiki Kataoka, and Takahiro Ishikawa.
\newblock {ChainerRL}: {A} deep reinforcement learning library.
\newblock \emph{Journal of Machine Learning Research}, 22\penalty0 (77):\penalty0 1--14, 2021.

\bibitem[Gomez et~al.(2024)Gomez, Bowling, and Machado]{gomezproper}
Diego Gomez, Michael Bowling, and Marlos~C. Machado.
\newblock Proper {Laplacian} representation learning.
\newblock In \emph{International Conference on Learning Representations}, 2024.

\bibitem[Harb et~al.(2018)Harb, Bacon, Klissarov, and Precup]{deliberation}
Jean Harb, Pierre-Luc Bacon, Martin Klissarov, and Doina Precup.
\newblock When waiting is not an option: Learning options with a deliberation cost.
\newblock In \emph{AAAI Conference on Artificial Intelligence}, 2018.

\bibitem[Jinnai et~al.(2019)Jinnai, Park, Abel, and Konidaris]{pmlr-v97-jinnai19b}
Yuu Jinnai, Jee~Won Park, David Abel, and George Konidaris.
\newblock Discovering options for exploration by minimizing cover time.
\newblock In \emph{International Conference on Machine Learning}, 2019.

\bibitem[Jinnai et~al.(2020)Jinnai, Park, Machado, and Konidaris]{jinnai2020exploration}
Yuu Jinnai, Jee~Won Park, Marlos~C. Machado, and George Konidaris.
\newblock Exploration in reinforcement learning with deep covering options.
\newblock In \emph{International Conference on Learning Representations}, 2020.

\bibitem[Jong et~al.(2008)Jong, Hester, and Stone]{Jong08}
Nicholas~K. Jong, Todd Hester, and Peter Stone.
\newblock The utility of temporal abstraction in reinforcement learning.
\newblock In \emph{International Joint Conference on Autonomous Agents and Multiagent Systems}, 2008.

\bibitem[Kingma \& Ba(2015)Kingma and Ba]{KingmaB14}
Diederik~P. Kingma and Jimmy Ba.
\newblock Adam: {A} method for stochastic optimization.
\newblock In \emph{International Conference on Learning Representations}, 2015.

\bibitem[Klissarov \& Machado(2023)Klissarov and Machado]{klissarov2023deep}
Martin Klissarov and Marlos~C. Machado.
\newblock Deep {Laplacian}-based options for temporally-extended exploration.
\newblock In \emph{International Conference on Machine Learning}, 2023.

\bibitem[Klissarov et~al.(2025)Klissarov, Bagaria, Luo, Konidaris, Precup, and Machado]{Klissarov2025}
Martin Klissarov, Akhil Bagaria, Ziyan Luo, George~Dimitri Konidaris, Doina Precup, and Marlos~C. Machado.
\newblock Discovering temporal structure: {A}n overview of hierarchical reinforcement learning.
\newblock \emph{CoRR}, abs/2506.14045, 2025.

\bibitem[Konidaris \& Barto(2007)Konidaris and Barto]{Konidaris2007BuildingPO}
George Konidaris and Andrew Barto.
\newblock Building portable options: Skill transfer in reinforcement learning.
\newblock In \emph{International Joint Conference on Artificial Intelligence}, 2007.

\bibitem[Konidaris \& Barto(2009)Konidaris and Barto]{skillchaining}
George Konidaris and Andrew Barto.
\newblock Skill discovery in continuous reinforcement learning domains using skill chaining.
\newblock In \emph{Neural Information Processing Systems}, 2009.

\bibitem[Kulkarni et~al.(2016)Kulkarni, Narasimhan, Saeedi, and Tenenbaum]{kulkarni2016hierarchical}
Tejas~D. Kulkarni, Karthik Narasimhan, Ardavan Saeedi, and Josh Tenenbaum.
\newblock Hierarchical deep reinforcement learning: Integrating temporal abstraction and intrinsic motivation.
\newblock In \emph{Neural Information Processing Systems}, 2016.

\bibitem[Levy et~al.(2019)Levy, Konidaris, Platt, and Saenko]{levy2019learning}
Andrew Levy, George Konidaris, Robert Platt, and Kate Saenko.
\newblock Learning multi-level hierarchies with hindsight.
\newblock In \emph{International Conference on Learning Representations}, 2019.

\bibitem[Liu et~al.(2017)Liu, Machado, Tesauro, and Campbell]{liu2017eigenoption}
Miao Liu, Marlos~C. Machado, Gerald Tesauro, and Murray Campbell.
\newblock The eigenoption-critic framework.
\newblock \emph{CoRR}, abs/1712.04065, 2017.

\bibitem[Machado(2025)]{Machado25}
Marlos~C. Machado.
\newblock Representation-driven option discovery in reinforcement learning.
\newblock In \emph{{AAAI Conference on Artificial Intelligence}}, 2025.

\bibitem[Machado \& Bowling(2016)Machado and Bowling]{machado2016learning}
Marlos~C. Machado and Michael Bowling.
\newblock Learning purposeful behaviour in the absence of rewards.
\newblock \emph{CoRR}, abs/1605.07700, 2016.

\bibitem[Machado et~al.(2017)Machado, Bellemare, and Bowling]{machado2017laplacian}
Marlos~C. Machado, Marc~G. Bellemare, and Michael Bowling.
\newblock A {Laplacian} framework for option discovery in reinforcement learning.
\newblock In \emph{International Conference on Machine Learning}, 2017.

\bibitem[Machado et~al.(2018)Machado, Rosenbaum, Guo, Liu, Tesauro, and Campbell]{machado2018eigenoption}
Marlos~C. Machado, Clemens Rosenbaum, Xiaoxiao Guo, Miao Liu, Gerald Tesauro, and Murray Campbell.
\newblock Eigenoption discovery through the deep successor representation.
\newblock In \emph{International Conference on Learning Representations}, 2018.

\bibitem[Machado et~al.(2023)Machado, Barreto, Precup, and Bowling]{machado2023temporal}
Marlos~C. Machado, Andre Barreto, Doina Precup, and Michael Bowling.
\newblock Temporal abstraction in reinforcement learning with the successor representation.
\newblock \emph{Journal of Machine Learning Research}, 24\penalty0 (80):\penalty0 1--69, 2023.

\bibitem[McGovern \& Barto(2001{\natexlab{a}})McGovern and Barto]{mcgovern2001accelerating}
Amy McGovern and Andrew Barto.
\newblock Accelerating reinforcement learning through the discovery of useful subgoals.
\newblock In \emph{International Symposium on Artificial Intelligence, Robotics, and Automation in Space}, 2001{\natexlab{a}}.

\bibitem[McGovern \& Barto(2001{\natexlab{b}})McGovern and Barto]{mcgovern2001automatic}
Amy McGovern and Andrew Barto.
\newblock Automatic discovery of subgoals in reinforcement learning using diverse density.
\newblock In \emph{International Conference on Machine Learning}, 2001{\natexlab{b}}.

\bibitem[Mnih et~al.(2015)Mnih, Kavukcuoglu, Silver, Rusu, Veness, Bellemare, Graves, Riedmiller, Fidjeland, Ostrovski, Petersen, Beattie, Sadik, Antonoglou, King, Kumaran, Wierstra, Legg, and Hassabis]{Mnih2015}
Volodymyr Mnih, Koray Kavukcuoglu, David Silver, Andrei~A. Rusu, Joel Veness, Marc~G. Bellemare, Alex Graves, Martin Riedmiller, Andreas~K. Fidjeland, Georg Ostrovski, Stig Petersen, Charles Beattie, Amir Sadik, Ioannis Antonoglou, Helen King, Dharshan Kumaran, Daan Wierstra, Shane Legg, and Demis Hassabis.
\newblock Human-level control through deep reinforcement learning.
\newblock \emph{Nature}, 518\penalty0 (7540):\penalty0 529--533, 2015.

\bibitem[Ouyang et~al.(2022)Ouyang, Wu, Jiang, Almeida, Wainwright, Mishkin, Zhang, Agarwal, Slama, Ray, Schulman, Hilton, Kelton, Miller, Simens, Askell, Welinder, Christiano, Leike, and Lowe]{Ouyang22}
Long Ouyang, Jeffrey Wu, Xu~Jiang, Diogo Almeida, Carroll~L. Wainwright, Pamela Mishkin, Chong Zhang, Sandhini Agarwal, Katarina Slama, Alex Ray, John Schulman, Jacob Hilton, Fraser Kelton, Luke Miller, Maddie Simens, Amanda Askell, Peter Welinder, Paul~F. Christiano, Jan Leike, and Ryan Lowe.
\newblock Training language models to follow instructions with human feedback.
\newblock In \emph{Neural Information Processing Systems}, 2022.

\bibitem[Piray \& Daw(2021)Piray and Daw]{Piray2021}
Payam Piray and Nathaniel~D. Daw.
\newblock Linear reinforcement learning in planning, grid fields, and cognitive control.
\newblock \emph{Nature Communications}, 12\penalty0 (1):\penalty0 4942, 2021.

\bibitem[Ramesh et~al.(2019)Ramesh, Tomar, and Ravindran]{ramesh2019successor}
Rahul Ramesh, Manan Tomar, and Balaraman Ravindran.
\newblock Successor options: An option discovery framework for reinforcement learning.
\newblock In \emph{International Joint Conference on Artificial Intelligence}, 2019.

\bibitem[Silver et~al.(2016)Silver, Huang, Maddison, Guez, Sifre, van~den Driessche, Schrittwieser, Antonoglou, Panneershelvam, Lanctot, Dieleman, Grewe, Nham, Kalchbrenner, Sutskever, Lillicrap, Leach, Kavukcuoglu, Graepel, and Hassabis]{Silver2016}
David Silver, Aja Huang, Chris~J. Maddison, Arthur Guez, Laurent Sifre, George van~den Driessche, Julian Schrittwieser, Ioannis Antonoglou, Veda Panneershelvam, Marc Lanctot, Sander Dieleman, Dominik Grewe, John Nham, Nal Kalchbrenner, Ilya Sutskever, Timothy Lillicrap, Madeleine Leach, Koray Kavukcuoglu, Thore Graepel, and Demis Hassabis.
\newblock Mastering the game of {Go} with deep neural networks and tree search.
\newblock \emph{Nature}, 529\penalty0 (7587):\penalty0 484--489, 2016.

\bibitem[Silver et~al.(2018)Silver, Hubert, Schrittwieser, Antonoglou, Lai, Guez, Lanctot, Sifre, Kumaran, Graepel, Lillicrap, Simonyan, and Hassabis]{silver2018general}
David Silver, Thomas Hubert, Julian Schrittwieser, Ioannis Antonoglou, Matthew Lai, Arthur Guez, Marc Lanctot, Laurent Sifre, Dharshan Kumaran, Thore Graepel, Timothy Lillicrap, Karen Simonyan, and Demis Hassabis.
\newblock A general reinforcement learning algorithm that masters chess, shogi, and go through self-play.
\newblock \emph{Science}, 362\penalty0 (6419):\penalty0 1140--1144, 2018.

\bibitem[{\c{S}}im{\c{s}}ek \& Barto(2004){\c{S}}im{\c{s}}ek and Barto]{csimcsek2004using}
{\"O}zg{\"u}r {\c{S}}im{\c{s}}ek and Andrew~G Barto.
\newblock Using relative novelty to identify useful temporal abstractions in reinforcement learning.
\newblock In \emph{International Conference on Machine Learning}, 2004.

\bibitem[Solway et~al.(2014)Solway, Diuk, C\'ordova, Yee, Barto, Niv, and Botvinick]{Solway14}
Alec Solway, Carlos Diuk, Natalia C\'ordova, Debbie Yee, Andrew Barto, Yael Niv, and Matthew~M. Botvinick.
\newblock {Optimal behavioral hierarchy}.
\newblock \emph{{PLOS Computational Biology}}, 10\penalty0 (8):\penalty0 1--10, 2014.

\bibitem[Sutton et~al.(1998)Sutton, Precup, and Singh]{sutton1998intra}
Richard~S. Sutton, Doina Precup, and Satinder Singh.
\newblock Intra-option learning about temporally abstract actions.
\newblock In \emph{International Conference on Machine Learning}, 1998.

\bibitem[Sutton et~al.(1999)Sutton, Precup, and Singh]{SUTTON1999181}
Richard~S. Sutton, Doina Precup, and Satinder Singh.
\newblock Between {MDPs} and semi-{MDPs}: A framework for temporal abstraction in reinforcement learning.
\newblock \emph{Artificial Intelligence}, 112\penalty0 (1):\penalty0 181--211, 1999.

\bibitem[Sutton et~al.(2023)Sutton, Machado, Holland, Szepesvari, Timbers, Tanner, and White]{sutton2023reward}
Richard~S. Sutton, Marlos~C. Machado, G.~Zacharias Holland, David Szepesvari, Finbarr Timbers, Brian Tanner, and Adam White.
\newblock Reward-respecting subtasks for model-based reinforcement learning.
\newblock \emph{Artificial Intelligence}, 324:\penalty0 104001, 2023.

\bibitem[Tse et~al.(2025)Tse, Chandrasekar, and Machado]{tse2025reward}
Hon~Tik Tse, Siddarth Chandrasekar, and Marlos~C. Machado.
\newblock Reward-aware proto-representations in reinforcement learning.
\newblock \emph{CoRR}, abs/2505.16217, 2025.

\bibitem[Van~Hasselt et~al.(2016)Van~Hasselt, Guez, and Silver]{van2016deep}
Hado Van~Hasselt, Arthur Guez, and David Silver.
\newblock Deep reinforcement learning with double {Q}-learning.
\newblock In \emph{AAAI Conference on Artificial Intelligence}, 2016.

\bibitem[Vezhnevets et~al.(2017)Vezhnevets, Osindero, Schaul, Heess, Jaderberg, Silver, and Kavukcuoglu]{vezhnevets2017feudal}
Alexander~Sasha Vezhnevets, Simon Osindero, Tom Schaul, Nicolas Heess, Max Jaderberg, David Silver, and Koray Kavukcuoglu.
\newblock Feudal networks for hierarchical reinforcement learning.
\newblock In \emph{International Conference on Machine Learning}, 2017.

\bibitem[Vinyals et~al.(2019)Vinyals, Babuschkin, Czarnecki, Mathieu, Dudzik, Chung, Choi, Powell, Ewalds, Georgiev, Oh, Horgan, Kroiss, Danihelka, Huang, Sifre, Cai, Agapiou, Jaderberg, Vezhnevets, Leblond, Pohlen, Dalibard, Budden, Sulsky, Molloy, Paine, G{\"{u}}l{\c{c}}ehre, Wang, Pfaff, Wu, Ring, Yogatama, W{\"{u}}nsch, McKinney, Smith, Schaul, Lillicrap, Kavukcuoglu, Hassabis, Apps, and Silver]{Vinyals19}
Oriol Vinyals, Igor Babuschkin, Wojciech~M. Czarnecki, Micha{\"{e}}l Mathieu, Andrew Dudzik, Junyoung Chung, David~H. Choi, Richard Powell, Timo Ewalds, Petko Georgiev, Junhyuk Oh, Dan Horgan, Manuel Kroiss, Ivo Danihelka, Aja Huang, Laurent Sifre, Trevor Cai, John~P. Agapiou, Max Jaderberg, Alexander~Sasha Vezhnevets, R{\'{e}}mi Leblond, Tobias Pohlen, Valentin Dalibard, David Budden, Yury Sulsky, James Molloy, Tom~Le Paine, {\c{C}}aglar G{\"{u}}l{\c{c}}ehre, Ziyu Wang, Tobias Pfaff, Yuhuai Wu, Roman Ring, Dani Yogatama, Dario W{\"{u}}nsch, Katrina McKinney, Oliver Smith, Tom Schaul, Timothy~P. Lillicrap, Koray Kavukcuoglu, Demis Hassabis, Chris Apps, and David Silver.
\newblock Grandmaster level in {StarCraft} {II} using multi-agent reinforcement learning.
\newblock \emph{Nature}, 575\penalty0 (7782):\penalty0 350--354, 2019.

\bibitem[Wang et~al.(2021)Wang, Zhou, Zhang, Shao, Hooi, and Feng]{wang2021towards}
Kaixin Wang, Kuangqi Zhou, Qixin Zhang, Jie Shao, Bryan Hooi, and Jiashi Feng.
\newblock Towards better {Laplacian} representation in reinforcement learning with generalized graph drawing.
\newblock In \emph{International Conference on Machine Learning}, 2021.

\bibitem[Watkins \& Dayan(1992)Watkins and Dayan]{Watkins1992}
Christopher J. C.~H. Watkins and Peter Dayan.
\newblock Q-learning.
\newblock \emph{Machine Learning}, 8\penalty0 (3):\penalty0 279--292, 1992.

\bibitem[Wu et~al.(2018)Wu, Tucker, and Nachum]{wulaplacian}
Yifan Wu, George Tucker, and Ofir Nachum.
\newblock The {Laplacian} in {RL}: Learning representations with efficient approximations.
\newblock In \emph{International Conference on Learning Representations}, 2018.

\end{thebibliography}
\bibliographystyle{rlj}

\beginSupplementaryMaterials
\section{VACE} \label{appendix:VACE}
In this section, we show the pseudocode and learning curves for VACE compared against baselines in the four rooms and nine rooms environments. 

VACE discovers options every $N_{steps}$ steps and adds them to its action space. This modifies its state visitation distribution compared to CEO, affecting options discovered in the future. It also learns option-values for the options discovered, allowing it to exploit options that lead to high reward. The pseudocode is shown in Algorithm \ref{alg:VACE}.

\vspace{12pt}
\begin{algorithm}[H] 
\caption{Value-Aware Covering Eigenoptions (VACE)}\label{alg:VACE}
\begin{algorithmic}
\Require $\eta, \alpha_o, \alpha \,;$ \Comment{Step-sizes for the SR and option policies and intra-option Q-learning} \\
        $\quad\quad\gamma_o, \gamma \,; $ \Comment{Discount factor for option policies and intra-option Q-learning}\\
        $\quad\quad N_{\text{steps}} \,;$ \Comment{Number of samples per option created} \\
        $\quad\quad N_{\text{sweeps}} \,;$ \Comment{Number of sweeps to learn SR from dataset} \\
        $\quad\quad N_{\text{iter}} \,;$ \Comment{Number of iterations of VACE} \\
        $\quad\quad\epsilon$ \Comment {Epsilon for epsilon-greedy exploration}

\State $\mathcal{D} \leftarrow \emptyset$ 
\State $\Omega \leftarrow \emptyset$
\State $Q \leftarrow |\mathcal{S}| \times |\mathcal{A}|$ matrix
\
\For{$i=1 \textbf{ to } N_{\text{iter}}$}
\For{$j=1 \textbf{ to } N_{\text{steps}}$}
    \State $s \leftarrow $ current state
    \If {$\text{Uniform}(0, 1) < \epsilon$}
        \State Choose $a$ randomly from $\mathcal{A} \cup \Omega$
    \Else
        \State $a \leftarrow \argmax_a Q(s, a)$; break ties randomly
    \EndIf
    \State $\mathcal{D}_o \leftarrow \emptyset$
    \If{$a \in \mathcal{A}$}
        \State Take action $a$, receive $r, s'$
        \State Append $(s, a, r, s')$ to $\mathcal{D}$ and $\mathcal{D}_o$
        \State $\text{UpdateOptionValues}(a, \mathcal{D}_o, s')$ (see Algorithm \ref{alg:optionvalues})
        \Else
        \While {option has not terminated}
            \State Take action $a_o$ according to option policy, receive $s', r$
            \State Append $(s, a_o, r, s')$ to $\mathcal{D}$ and $\mathcal{D}_o$
        \EndWhile
        \State $\text{UpdateOptionValues}(a, \mathcal{D}_o, s')$ (see Algorithm \ref{alg:optionvalues})
    \EndIf
\EndFor
\EndFor

\State Learn SR by applying Equation \ref{eqn:onlineSR} sweeping through $\mathcal{D}$ $N_\text{sweeps}$ times.
\State Get top eigenvector of SR and create intrinsic reward function using Equation \ref{eqn:intrinsic_reward}.
\State Learn an option policy to maximize intrinsic reward using Q-learning.
\State Define termination states as all states with $Q(s, a) \leq 0$.
\State Append option to $\Omega$, append column of zeros to $Q$.

\end{algorithmic}
\end{algorithm}

\begin{algorithm}[t] 
\caption{Update Option Values}\label{alg:optionvalues}
\begin{algorithmic}
    \Require $o$ \Comment{Option} \\
             $\quad\quad\mathcal{D}_o$ \Comment{Sequence of samples collected while taking the option}\\
             $\quad\quad s_\text{next}$ \Comment{State the agent is in after taking option $o$}
    \State \text{return} $\leftarrow 0$
    \For {$(s, a, r, s') \textbf{ in } \text{reverse}(\mathcal{D}_o)$}
    \State $\text{return} \leftarrow r + \gamma \cdot \text{return}$
    \State $Q(s, o) \leftarrow Q (s, o) + \alpha \left[\left(\text{return} + \gamma \argmax_{o' \in \mathcal{O}}Q(s_\text{next}, o')\right) - Q(s, o)\right]$ 
    \For {$\hat{o} \textbf{ in } \Omega \setminus \{o\}$}
        \If{$\pi_{\hat{o}}(s) = a$}
        \State $Q(s, \hat{o}) \leftarrow Q (s, \hat{o}) + \alpha \left[\left(r + \gamma U(s', \hat{o})\right) - Q(s, \hat{o})\right]$
        \EndIf
    \EndFor
    \EndFor

\end{algorithmic}
\end{algorithm}

\begin{figure}[t]
    \centering
    \includegraphics[width=\linewidth]{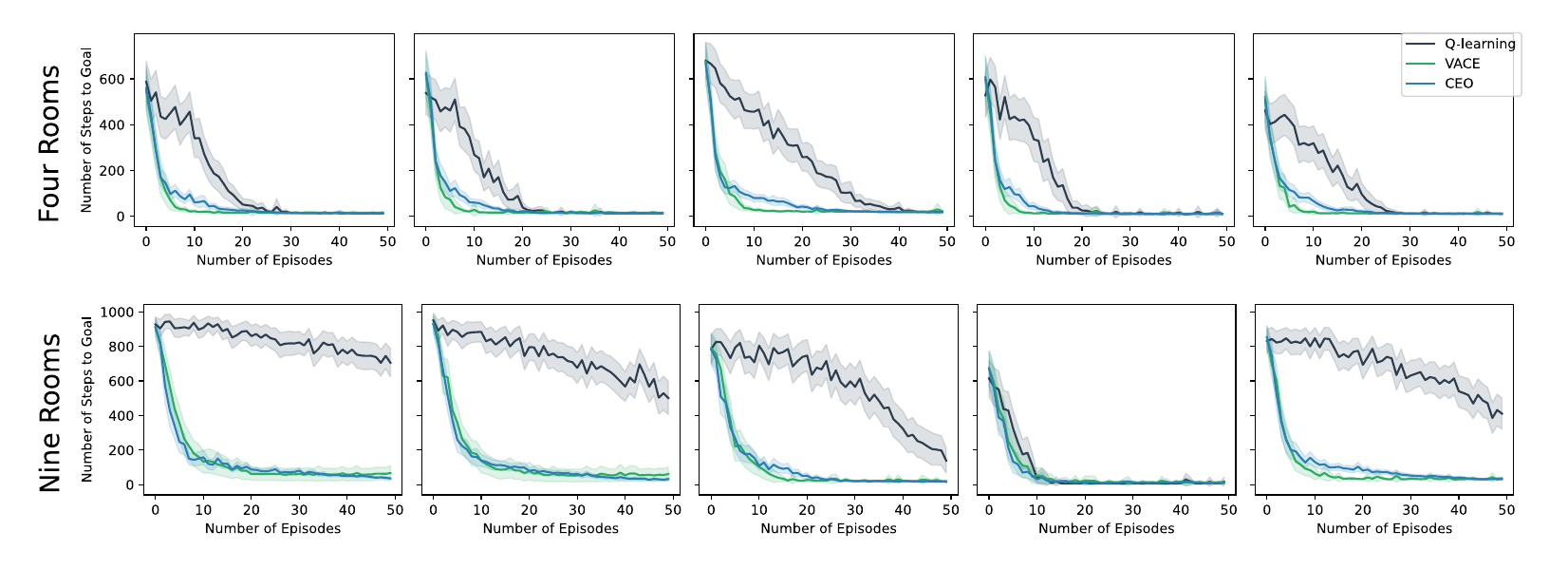}
    \caption{Performance of VACE and baselines when discovering options online every 1000 time steps. Environment configurations are the same as in Fig. \ref{fig:tabVAEO}. We evaluate all algorithms on 100 independent runs, and the shaded region represents a $99\%$ confidence interval.}
    \label{fig:ROD}
\end{figure}
\begin{figure}[H]
    \centering
    \includegraphics[width=\linewidth]{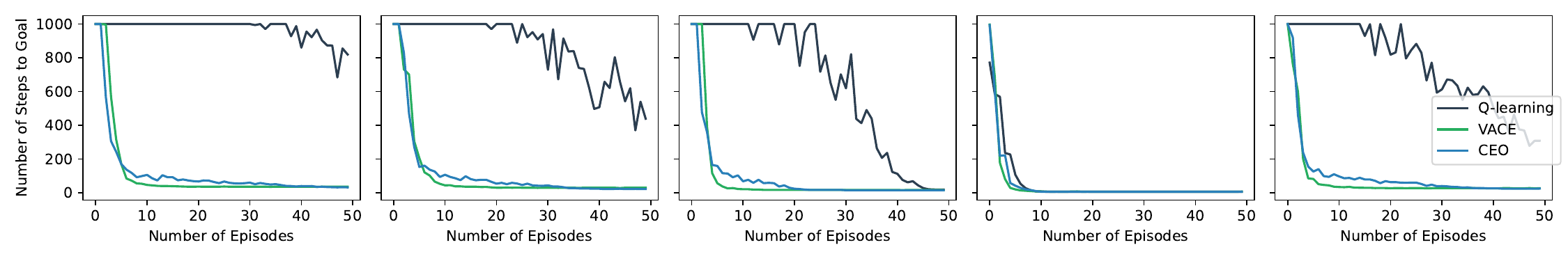}
    \caption{Median over 100 independent runs for VACE and baselines in the nine rooms environment.}
    \label{fig:RODMedian}
\end{figure}

\vspace{20pt}
Figure \ref{fig:ROD} shows the learning curves of VACE against the baselines of CEO and Q-learning. VACE and CEO discover options every 1000 steps. VACE shows slightly faster learning than CEO in the four rooms environment but shows similar performance in the nine rooms environment. This is because, in the nine rooms environment, VACE performs better than CEO in most runs, however, it also has a few runs where it performs significantly worse than CEO. This is further explained in Figure \ref{fig:valueprop}, which shows the state visitation distributions and value propagation during such runs. 
Figure \ref{fig:RODMedian} shows median curves in the nine rooms environments and confirms that VACE's median run is slightly better than CEO's median run in the nine rooms environment.\\

\vspace{-1pt}

\section{DVAEO} \label{appendix:DVAEO}

In this section, we show the hierarchical DQN architecture that we use along with the learning curves for DVAEO (with approximated option policies) compared against baselines in the four rooms and nine rooms environments.

\begin{figure}[t]
    \begin{center}
        \includegraphics[width=0.8\linewidth]{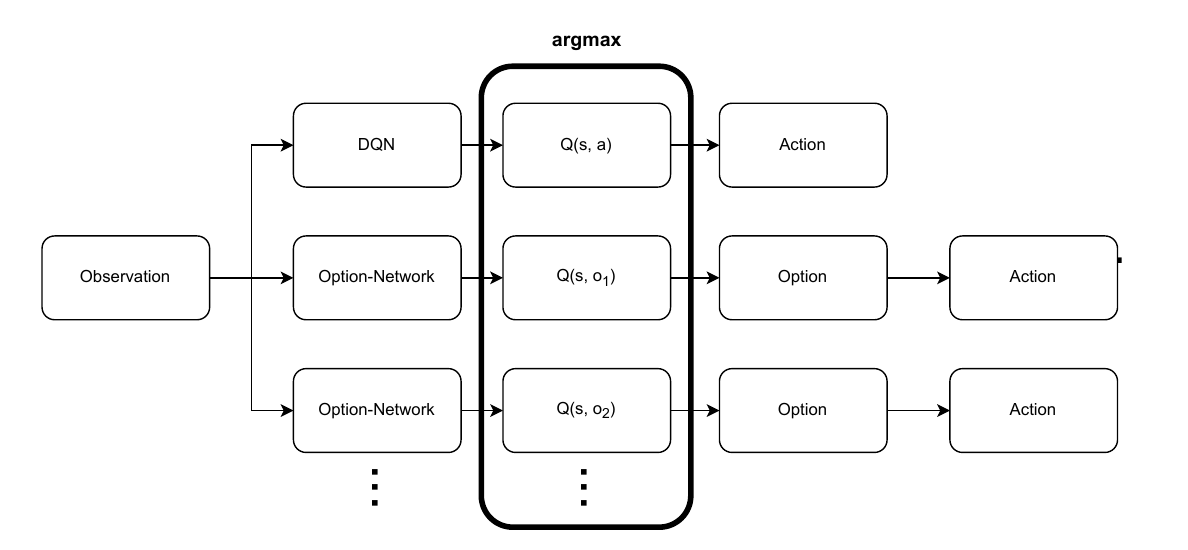}
    \end{center}
    \caption{Hierarchical DQN architecture. We include a separate network to predict the value of each option. If an option has the highest option-value, its respective option policy will choose an action. }
    \label{fig:arch}
\end{figure}

Figure \ref{fig:arch} shows the hierachical DQN architecture that we use. We use a separate network to predict option-values for each option. To select an action greedily, we take an argmax over the action-values from the action-value network and the option-values from the option-value networks.

\begin{figure}[t]
    \centering
    \includegraphics[width=\linewidth]{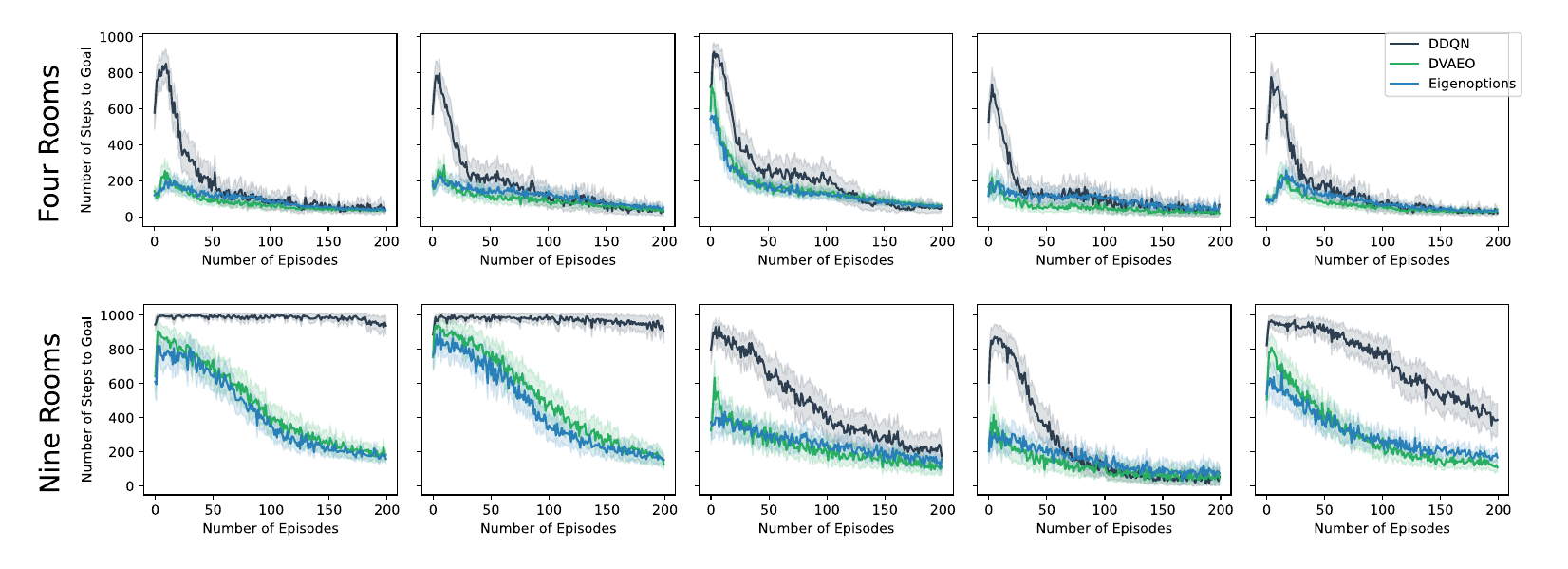}
    \caption{Performance of DVAEO (with approximated option policies) and baselines with pixel observations. Environment configurations are the same as in Fig.~\ref{fig:tabVAEO}. We use 6 eigenoptions in the four rooms environment and 24 eigenoptions in the nine rooms environment. We train all algorithms for 100 independent trials, and the shaded region represents a $99\%$ confidence interval.}
    \label{fig:FA}
\end{figure}

Figure \ref{fig:FA} shows the learning curves of DVAEO compared against the baselines of deep eigenoptions and DDQN with $\epsilon$-greedy exploration. DVAEO and deep eigenoptions both use eigenoption policies approximated using DDQN with termination after 10 steps.  

\section{Hyperparameters and Experimental Details}
We run all our experiments in modified versions of the Minigrid environment~\citep{MinigridMiniworld23} using Gym~\citep{gym}. The original Minigrid environment included actions to rotate the agent and move the agent. We use a simplified action set where the agent can move up, down, right and left.
For clarity, we outline the differences among algorithms introduced in each section in Table \ref{tab:algorithms}.
\begin{table}[htbp] 
    \caption{Disambiguation of algorithms introduced in each section of the paper. Approximated refers to non-linear approximation with neural networks. Online refers to the use of the ROD cycle to generate eigenoptions.}
    \label{tab:algorithms}
    \begin{center}
        \begin{tabular}{llllll}
            \multicolumn{1}{l}{\bf Algorithm}
            &\multicolumn{1}{l}{\bf Section}
            &\multicolumn{1}{l}{\bf Option Discovery}
            &\multicolumn{1}{l}{\bf Eigenvectors}
            &\multicolumn{1}{l}{\bf Option Policies}
            &\multicolumn{1}{l}{\bf Option Values}
            \\ \hline \\
            VAEO & \ref{sec:tabular} & Offline & Tabular & Tabular & Tabular  \\
            VACE & \ref{sec:tabularROD} & Online & Tabular & Tabular & Tabular  \\
            DVAEO & \ref{sec:FAtabular} & Offline & Tabular & Tabular &  Approximated\\
            DVAEO & \ref{sec:approx} & Offline & Tabular & Approximated &  Approximated\\
        \end{tabular}
    \end{center}
    \label{tab:exampleTable}
\end{table}

For the algorithms in Figure \ref{fig:tabVAEO}, we use $\epsilon = 0.05$, $\gamma, \gamma_o=0.99$, and $\alpha, \alpha_o=0.1$ where $\gamma_o$ and $\alpha_o$ correspond to the discount factor and step size in the intra-option updates. We conducted a hyperparameter sweep on the number of eigenoptions in each environment, leading to the choice of six eigenoptions in the four rooms domain and 24 eigenoptions in the nine rooms domain. We swept over $N_{\text{options}} = \{2, 4, 6, 8, 16, 24, 32\}$ in both domains. To learn eigenoption policies, we use Q-learning with $\gamma=0.9$ and $\alpha=0.1$. We run Q-learning for 100 episodes of 1000 steps each and use the intrinsic reward function. We use all bottleneck options in both environments. In the four rooms domain there are a total of eight bottleneck options, while there are 24 bottleneck options in the nine rooms domain.

For the algorithms in Figure \ref{fig:tabCredit}, we use the same values as in Figure \ref{fig:tabVAEO}. Intra-option and action updates happen only in the training phase. However, options cannot be used in the training phase and all intra-option updates are made when primitive actions are taken.

For the algorithms in Figures \ref{fig:valueprop}, \ref{fig:ROD}, and \ref{fig:RODMedian}, we use $\epsilon=0.05$; $\eta, \alpha, \alpha_o = 0.1$; $\gamma, \gamma_o = 0.99$, and $N_\text{steps} = 1000$. We ran a hyperparameter sweep over $N_\text{steps} = \{100, 500, 1000, 10000\}$ in both the four rooms and nine rooms domains. To learn eigenoption policies, we store all samples in a dataset and sweep over the dataset $N_\text{sweeps}=100$ times using Q-learning with $\gamma=0.9$ and $\alpha=0.1$. \looseness=-1

For the algorithms in Figures \ref{fig:FATab} and \ref{fig:FA}, we use modified implementations from PFRL \citep{pfrl}. Pixel observations from the environments were of size $52 \times 52 \times 3$. For all algorithms, we use $\gamma=0.9$, an update interval of 1 step, a target update interval of 2000 steps, a replay buffer of size 20,000, and the Adam optimizer~\citep{KingmaB14} with a step size of $10^{-5}$. We use linear $\epsilon$ decay with $\epsilon=0.1$ decaying to $\epsilon=0.05$ over 180,000 steps. We also use six eigenoptions in the four rooms domain and 24 eigenoptions in the nine rooms domain just as we did with our tabular experiments. However, we did not perform a hyperparameter sweep over the number of options in the deep RL setting.

To learn eigenoption policies, we train a DDQN model, with $\epsilon=1$ and a replay buffer of size 100,000, on each eigenvector of the SR for 500 episodes of length 1000.  When eigenoption policies are being used in either deep eigenoptions or DVAEO, they are no longer updated and are used in evaluation mode.

For all deep RL algorithms (including eigenoption policies), we use a two-layer convolutional network with 32 channels each, a 3 $\times$ 3 kernel, and a stride of 2, followed by a fully connected layer with 256 nodes and then the output layer with 4 nodes (one for each action) in the action-value network and 1 node in the option-value network.

\end{document}